\documentclass{article}
\usepackage{kr}

\usepackage{times}
\usepackage{soul}
\usepackage{url}
\usepackage[hidelinks]{hyperref}
\usepackage[utf8]{inputenc}
\usepackage[small]{caption}
\usepackage{graphicx}
\usepackage{amsmath}
\usepackage{amsthm}
\usepackage{booktabs}
\usepackage{algorithm}
\usepackage{algorithmicx}
\usepackage{algpseudocode}

\usepackage{amssymb}
\usepackage{pifont}
\usepackage{multirow}
\usepackage{color}

\title{Hybrid Augmented Automated Graph Contrastive Learning}

\author{%
Yifu Chen\and
Qianqian Ren \and
Yong Liu\\
\affiliations
Heilongjiang University\\
\emails
2211892@s.hlju.edu.cn, renqianqian@hlju.edu.cn
}

\begin{document}

\maketitle

\begin{abstract}
  Graph augmentations are essential for graph contrastive learning. Most existing works use pre-defined random augmentations, which are usually unable to adapt to different input graphs and fail to consider the impact of different nodes and edges on graph semantics. To address this issue, we propose a framework called Hybrid Augmented Automated Graph Contrastive Learning (HAGCL). HAGCL consists of a feature-level learnable view generator and an edge-level learnable view generator. The view generators are end-to-end differentiable to learn the probability distribution of views conditioned on the input graph. It insures to learn the most semantically meaningful structure in terms of features and topology, respectively. Furthermore, we propose an improved joint training strategy, which can achieve better results than previous works without resorting to any weak label information in the downstream tasks and extensive evaluation of additional work.
\end{abstract}

\section{Introduction}
Contrastive learning has demonstrated promising performance in supervised methods for various fields. It leverages the principle of Mutual Information Maximization (InfoMax) \cite{linsker1988self} to maximize the correspondence between different view representations of the graph. To enhance the performance of contrastive learning,
many studies employ graph augmentation methods. For example, GRACE \cite{zhu2020deep} combines random node features masking and random edges dropping strategies to generate contrastive samples. GraphCL \cite{you2020graph} provides a variety of augmentation strategies to adapt the most suitable augmentation approach for the input data.

\par However, graphs are typically abstract representations of raw data with diverse properties, which makes it challenging to provide a globally effective augmentation scheme \cite{you2020graph}. Moreover, random graph augmentations, such as uniform edge drop, may also result in the destruction of crucial graph structures, thereby affecting the graph semantics. GCA \cite{zhu2021graph} further considers the differences in the influence of nodes and edges during data augmentation, respectively dropping unimportant node features and edges to ensure the semantics of a graph based on centrality. Nonetheless, centrality as prior domain knowledge is task-dependent and not learnable.

\par In order to enable data augmentation to adapt to input data, it is worthwhile to design a graph data augmentation scheme that can be automatically optimized based on the input data. AutoGCL \cite{yin2022autogcl} applies InfoMin \cite{tian2020makes} to the graph data, utilizing learnable view generators to learn the probability distribution of contrasting views conditioned on the input graph. However, this approach cannot guarantee the removal of redundant information in unsupervised tasks.InfoMin is more suitable for semi-supervised tasks. Additionally, AutoGCL only selects node-level subgraphs and feature masking as augmentations, and fails to consider the impact of topology.

\par Some studies have raised concerns about InfoMax, as it may encourage the encoder to capture redundant information that is irrelevant to downstream tasks \cite{tschannen2019mutual}. 
In contrast to InfoMax, AD-GCL \cite{suresh2021adversarial} leverages an adversarial strategy to encourage the minimization of sufficient information of the original graph data when using GIB \cite{wu2020graph} in unsupervised learning. When graph data augmentation is highly aggressive, it aims to train the encoder to maximize the mutual information between the augmented graph and the original graph. Similarly, the data augmentation is learnable and can learn the probability of edge deletion. AD-GGL has better theoretical support in unsupervised tasks, as it minimizes the mutual information between the view and the original graph to remove redundancy, and then maximizes the mutual information between the view and the original graph to obtain an optimal representation.

\par AD-GCL did not take into account the impact of node redundancy information, and it was not fully automated as the method still required regularization terms to replace the augmentation rate to ensure semantic preservation. Moreover, when maximizing mutual information, AD-GCL only optimized the encoder without optimizing the graph data augmentation. Merely constraining graph data augmentation by minimizing mutual information may not be sufficient and could lead to excessively blind optimization.
This indicates that joint training strategies can be used to further ensure semantic preservation and optimize the search space of data augmentation. Another benefit of this approach is that regularization is not required, as regularization parameters often require extensive evaluation of input data, which entails significant additional work.
 
\par Targeting the shortcomings of existing studies, we propose a novel learnable view generation framework, named Hybrid Augmented Automated Graph Contrastive Learning (HAGCL).
Specifically, we first design a GNN encoder to maximize the mutual information between the original graph and the two view representations. Then, a GNN-based learnable feature masking view generator and a GNN-based learnable edge drop view generator are designed. Both are end-to-end differentiable due to the gumble-softmax trick \cite{jang2016categorical}.

\par Additionally, different augmentation methods naturally provide sufficient augmentation variances for the contrastive samples. We expect the view generators to learn the probability distribution of the augmented view, limited by the input graph, not only by minimizing mutual information but also by maximizing mutual information, to reduce redundancy and preserve semantic information from the perspective of node features and topology.
In summary, the main contributions of this paper are as follows:
\begin{itemize}
    \item We propose a novel framework for graph contrastive learning, learnable graph view generators driven by an improved joint training strategy to generate optimal contrastive views from two perspectives of node features and topology. Compared to previous models, our approach demonstrates superior effectiveness and efficiency.
    \item We design an improved joint training strategy and incorporate it into our framework, which allows our model to avoid introducing regularization parameters and augmentation rates. These parameters require weak information from downstream tasks and extensive evaluation efforts. Our approach has truly achieved automation in unsupervised learning.
    \item We extensively evaluate the proposed method on various datasets in unsupervised tasks, and the results demonstrate that HAGCL outperforms the state-of-the-art methods in terms of performance.
\end{itemize}

\section{Related Works}
\subsection{Graph Neural Network}
This work mainly deals with graph classification tasks using graph neural networks as encoders. Define the attribute graph as $g=(V,E)$, where $V$ is the node set, and for any $v \in V$, $x_v$ is represented as the node feature, $E$ is an edge set, and we also consider additional edge attributes. For any $e \in E$, $x_e$ is represented as an edge feature. We denote the set of neighbors of a node $v$ as $\mathcal{N}(v)$.
GNN obtains node embedding $h_v$ by aggregating neighbor node features $x_v$, and each layer of GNN is used as an iteration of aggregation, then the node embedding after the $k$ layer will aggregate the features of its $k$ hop neighbors. We denote the hidden state of nodes after the $k$-th layer as $\boldsymbol{h}_v^{(k)}$, where $\boldsymbol{h}_v^{(0)}=x_v$ represents the original input node features.
The $k$-th layer of GNN can be expressed as
\setlength{\abovedisplayskip}{0pt}
\setlength{\belowdisplayskip}{0pt}
\begin{equation}
\boldsymbol{a}_v^{(k)}=\operatorname{AGGREGATE}^{(k)}\left(\left\{\boldsymbol{h}_u^{(k-1)}: u \in \mathcal{N}(v)\right\}\right)
\end{equation}
\setlength{\abovedisplayskip}{0pt}
\setlength{\belowdisplayskip}{0pt}
\begin{equation}
\boldsymbol{h}_v^{(k)}=\operatorname{COMBINE}{ }^{(k)}\left(\boldsymbol{h}_v^{(k-1)},\boldsymbol{a}_v^{(k)}\right)
\end{equation}
Among them, $\operatorname{AGGREGATE}$ is a learnable function that maps the neighbors of node $v$ to an aggregation vector $\boldsymbol{a}_v$, and $\operatorname{COMBINE}$ is also a learnable function, which maps the neighbor aggregation vector $\boldsymbol{a}_v$ of node $v$ and the current representation of $v$ to the update vector of the current layer of $v$.

For the graph classification task, it is necessary to pool the node embeddings into graph embeddings, and further map the graph embeddings to the graph embedding dimensions we need through the MLP layer. can be expressed as
\begin{equation}
\boldsymbol{z}_g=\operatorname{MLP}\left(\operatorname{POOL}\left(\left\{\boldsymbol{h}_u^{(k)}: u \in \mathcal{V}\right\}\right)\right)
\end{equation}

\subsection{Graph Contrastive Learning}
Contrastive learning was originally applied in the computer vision domain. Due to human prior knowledge of image semantics, image semantics can be effectively preserved by data augmentation, and this unique perspective can serve as natural contrastive samples. In contrastive learning, the InfoMax method is used to train the GNN encoder for graph representation learning. By maximizing mutual information, views of the same input are pulled together in the representation space, while views of different inputs are pulled apart. Unlike images, graph data is more abstract and complex, and human prior knowledge of graph semantics is insufficient. Therefore, obtaining more suitable contrastive samples in graph contrastive learning presents a challenge. The contrastive samples of graph contrastive learning are mainly reflected in two aspects. Part of the work uses different parts of the graph as contrastive samples. The contrastive samples of DGI\cite{velickovic2019deep} are the node-level representation and graph representation of the same graph, GMI\cite{peng2020graph} is between nodes, SUBG-CON\cite{jiao2020sub} is between the node and the subgraph.
At present, most of the work \cite{he2020momentum,grill2020bootstrap,qiu2020gcc} utilizes data augmentation to generate views as contrastive samples.

\par GraphCL\cite{you2020graph} has conducted extensive experiments and research on different types of data augmentation combinations, including node drop, edge perturbation, subgraph and feature masking, to select the optimal solution for specific downstream tasks.
GCA\cite{zhu2021graph} uses node feature masking and edge drop as data augmentation, and utilizes the domain knowledge of network centrality to adaptively optimize data augmentation.
These tasks either require domain knowledge to be exploited or extensive evaluations to be conducted, which are not learnable.

\subsection{Learnable Data Augmentation}
The optimal data augmentation method is task-dependent, and pre-designed data augmentation may not meet the requirements of various unknown data. Currently, researchers are exploring learnable data augmentation methods. JOAO \cite{you2021graph} learned combinations of different augmentation methods, but the augmentation itself was not learnable. With the emergence of the Gumbel-Softmax tricks, data augmentation can be trained end-to-end. AD-GCL \cite{suresh2021adversarial} first implemented automatic edge dropping, but still required a regularization term as an augmentation rate to retain the semantic lower bound. The regularization hyperparameters need to be extensively evaluated using weak labels in downstream tasks, and AD-GCL did not truly achieve automation. AutoGCL \cite{yin2022autogcl} still requires labels and is not suitable for unsupervised learning, but more suitable for semi-supervised learning.
For unsupervised tasks, AD-GCL believes that data augmentation should be as aggressive as possible to achieve redundancy reduction, but this requires retaining sufficient semantics. This part of semantics is manually adjusted and requires weak labels in downstream tasks. Meanwhile, AD-GCL only considered edges and did not consider redundancy on nodes. AutoGCL hopes to retain semantics as much as possible, but it did not consider redundancy. If no augmentation is performed on the original data, comparing two identical views will preserve all semantics, but this is meaningless because it will make GNN unable to distinguish between redundancy and semantics. AutoGCL also ignored edge-level augmentation.
Currently, there are still incomplete issues regarding the effectiveness and automation of learnable data augmentation in research. Our research focuses on how to automatically obtain more effective augmentation views. We augment the two basic structures of graph data, node features, and topological structure, respectively, in order to remove redundancy as much as possible while retaining semantics. Combined with our joint training optimization strategy, these are all automatically completed without any labels. As the different augmentation angles provide natural augmentation variance, this means that the views are different enough from each other. This also means that GNN can learn semantic commonalities from both node features and topology, rather than just one of the two.
We have summarized in Table \ref{tab3}.
\begin{table}
\centering
\scalebox{0.55}{
\begin{tabular}{llllll}
\hline
Moedels  & Topological & Node  & Automaticity & Remove Redundancy & Preserve Semantics\\
\hline
JOAO       & \makebox[0.2\linewidth][c]{\ding{51}}  & \makebox[0.1\linewidth][c]{\ding{51}} & & &    \\
AD-GCL     & \makebox[0.2\linewidth][c]{\ding{51}}  & &\makebox[0.2\linewidth][c]{\ding{51}} & \makebox[0.35\linewidth][c]{\ding{51}}&     \\
AutoGCL    &   & \makebox[0.1\linewidth][c]{\ding{51}} &\makebox[0.2\linewidth][c]{\ding{51}} & &\makebox[0.35\linewidth][c]{\ding{51}}    \\

HAGCL      & \makebox[0.2\linewidth][c]{\ding{51}}  & \makebox[0.1\linewidth][c]{\ding{51}} & \makebox[0.2\linewidth][c]{\ding{51}} & \makebox[0.35\linewidth][c]{\ding{51}}  & \makebox[0.35\linewidth][c]{\ding{51}}  \\
\hline
\end{tabular}
}
\caption{An overview of learnable data augmentation.Compared to previous works, HAGCL is more comprehensive.}
\label{tab3}
\end{table}

\section{Methodology}
\begin{figure}[htp]
    \centering
    \includegraphics[width=0.45\textwidth]{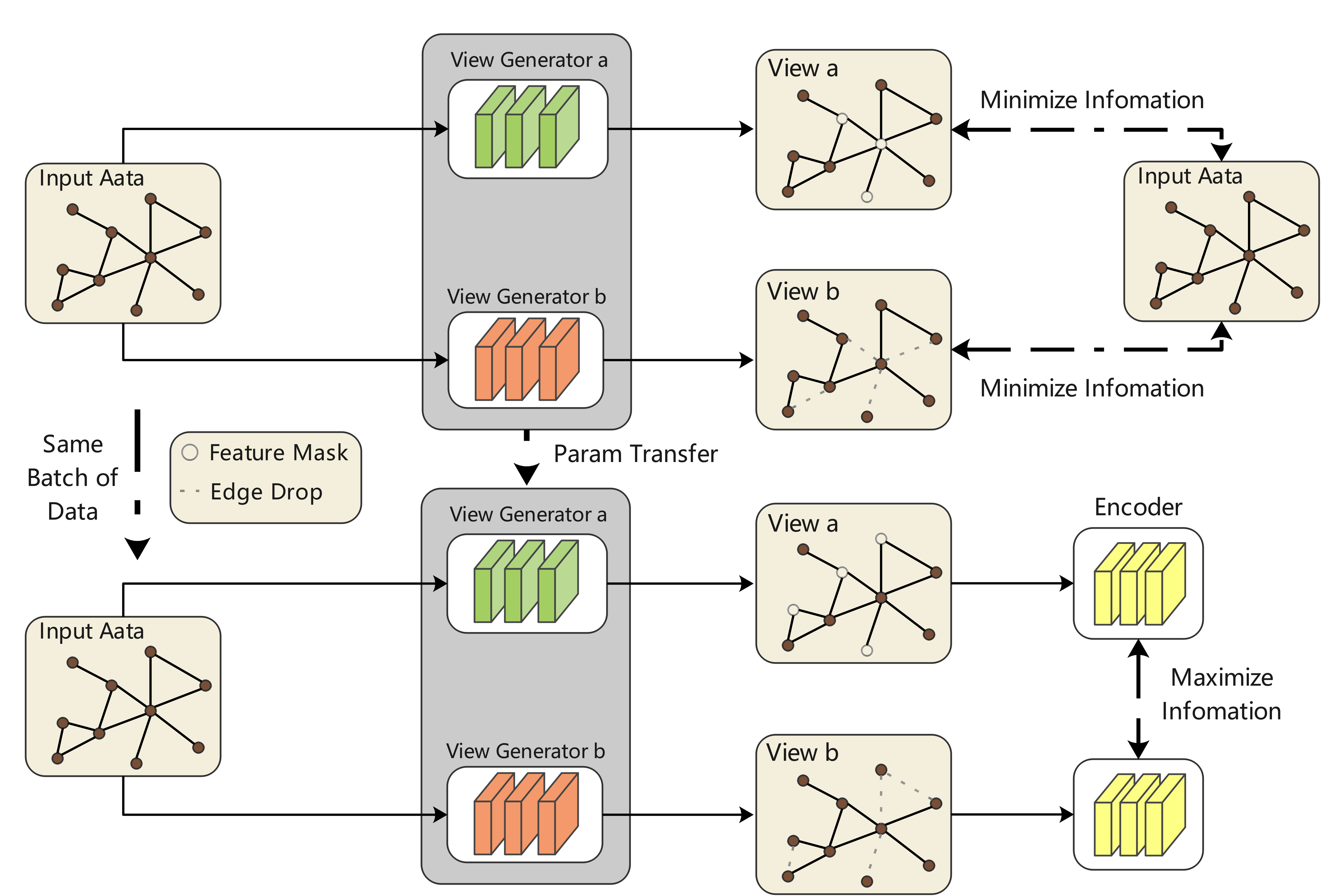}
    \caption{The HAGCL model comprises (1) an edge-level view generator and a feature-level view generator, both of which are composed of GNN augmentors. The view generators use joint optimization of InfoMax and mutual information minimization to generate views from the perspectives of features and topology, respectively. (2) A GNN encoder optimized by InfoMax is included in HAGCL to obtain view embeddings.}
    \label{fig1}
\end{figure}

In this section, we give an overview of the proposed HAGCL and elaborate on the designing detail.
Fig\ref{fig1} illustrates the overall architecture of the model. First, we elaborate a learnable view generator driven by a joint training strategy, which is able to learn optimal views of input graphs from both node features and topological perspectives. Second, we illustrate the uesed loss functions, which not only enable mutual information minimization and maximization, but also provide the basis for a joint training strategy. Finally, we detail the implementation of the joint training strategy.

\subsection{Learnable View Generator}
Pre-defined random augmentation (such as uniform edge drop) relies on the augmentation rate to ensure the lower bound of the semantic information carried by the augmented view. This lower bound only represents part of the graph semantics and cannot completely distinguish the important information of the graph data, which affects GNN downstream task performance.

To tackle above problem, this work designs a learnable data augmentation strategy driven by input data instead of random augmentation. Combined with our improved joint training strategy(Section 3.2), the learnable view generator can automatically generate an optimal view from the input data, which can preserve the semantic information of the input data and remove redundant information as much as possible. The design of the learnable view generator is as follows.

A view generator is typically divided into two types, that's feature level view generator and edge level view generator. These two types view generators are corresponding to two basic attributes of the graph. In particular, the feature-level view generator performs a masking operation on the node features in order to mask trivial node features; the edge-level view generator drops unimportant edges.
\begin{figure}
    \centering
    \includegraphics[width=0.47\textwidth]{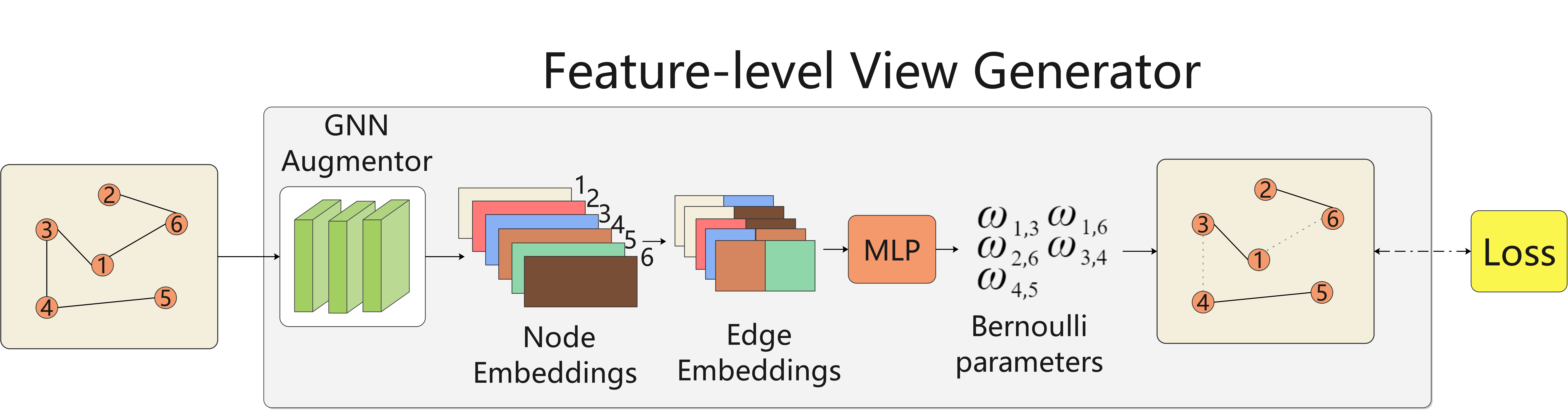}
    \caption{Architecture of the Edge-level View Generator. GNN is used to obtain node embeddings, which are then concated as edge embeddings based on topological relationships. Then, further projection and gumbel-softmax are used to convert them into Bernoulli parameters.}
    \label{fig2}
\end{figure}

Given the graph $G = (V, E)$, where $V=\{v_1, v_2,\cdots, v_N\}$ is the set of $N$ nodes, $E$ is the set of edges between two nodes and $X_v$ is the node feature vector.
We stack $K$ layers of GIN\cite{xu2018powerful} to obtain nodes embedding from the input graph. Let $\boldsymbol{h}_i^{(k)}(k=0,1,\cdots,K-1)$ represent the hidden state of node $i$ at layer $k$, $\mathcal{N}(i)$ represents all nodes connected to node $i$.
Then the GNN at $k-th$layer is formulated as
\begin{equation}
\boldsymbol{h}_i^{(k-1)}=\operatorname{COMBINE}{ }^{(k)}\left(\boldsymbol{h}_i^{(k-2)}, \boldsymbol{a}_i^{(k-1)}\right)
\end{equation}
\begin{equation}
\boldsymbol{a}_i^{(k)}=\operatorname{AGGREGATE}^{(k)}\left(\left\{\boldsymbol{h}_u^{(k-1)}: u \in \mathcal{N}(i)\right\}\right)
\end{equation}
We will elaborate on the proposed edge-level view generator and feature-level view generator in the rest of the section.
\subsubsection{Edge-Level View Generator}
Figure \ref{fig2} depicts our proposed feature-level view generator.
The edge-level view generator needs to perform edge drop. If node $i$ has an edge connection with node $j$, we concat the two node embedding as an edge embedding. The edge embedding between node $v_i$ and $v_j$ at $k-th$ layer is denoted as:
\begin{equation}
e_{ij}^{(k)}=\boldsymbol{a}_i^{(k)}||\boldsymbol{a}_j^{(k)}
\end{equation}
In order to make the view generator end-to-end differentiable, we compress the edge embedding vectors into Bernoulli parameters $\omega_{i,j}$ through MLP projection.
\begin{equation}
\omega_{i,j}=\operatorname{MLP}\left(e_{ij}^{(k)}\right)
\end{equation}

Next, we use the gumbel-softmax trick to reparameterize $\omega_{i,j}$ and project it to a random continuous variable $p_{i,j}$ between [0,1]. Each edge corresponds to a random variable as the probability of drop, that's
\begin{equation}
p_{i,j}=\operatorname{GumbelSoftmax}\left(\omega_{i,j}\right)
\end{equation}

Finally, the view representation of the edge-level view generator is obtained by sampling the random variable $p_{i,j}$.

\subsubsection{Feature-Level View Generator}
For the feature-level view generator, the node embeddings are feed into the MLP module without performing node connections when going through $k-th$ layer GIN. Considering the speciality of node masking, when reparameterizing the Bernoulli parameter with gumbel-softmax, we directly sample it as a one-hot Encoding. If the value associated with the node is 1, then the node features are retained, otherwise the masking operation is performed. The procedure is formulated as
\begin{equation}
\psi_{i}=\operatorname{MLP}\left(\boldsymbol{a}_i^{(k)}\right)
\end{equation}
\begin{equation}
p_{i}=\operatorname{GumbelSoftmax}\left(\psi_{i}\right)
\end{equation}
\subsection{Optimizated Strategies for Learnable View Generators}
The contrastive learning generally requires InfoMax, but it may capture too much redundancy irrelevant with the semantics of the input data. InfoMax cannot verify whether the positive sample is the minimum sufficient semantics of the corresponding input data. Although the semantic lower bound of the view is guaranteed by the limitation of the augment rate, if the redundant information is sufficient to identify positive and negative pairs, it may eventually cause GNN to learn these redundant information and ignore the real effective semantics.
The emergence of learnable view generators theoretically realizes end-to-end differentiability. However, how to train the view generator to generate the optimal augmented view in unsupervised tasks, that is, to retain enough semantics and remove redundancy as much as possible is still challenging.

For strategies to optimize data augmentation, AutoGCL believes that good positive sample views should maximize semantics and minimize mutual information between each other. However, semantics cannot be derived from labels in unsupervised tasks, so the requirement to reduce the similarity between positive views may harm semantics.
This also inspires us, in order to better optimize the view generators, it can be restricted by conditions besides InfoMax. The problem is how to find additional effective constraints in unsupervised tasks.
AD-GCL exploits the adversarial strategy without using labels. The data augmentation process minimizes the mutual information between the original graph and the view instead of Infomax to optimize the view generators.In the contrastive stage, InfoMax is used to maximize the mutual information of the original graph and the augmented view to train the GNN encoder. This adversarial strategy expects that even if the augmentation strategy is aggressive enough, the GNN encoder can still capture the remaining information.

To further optimize the learnable view generators, combining the advantages of joint training and adversariality, we propose a strategy to jointly and alternately train the view generator with mutual information minimization and InfoMax.

\subsubsection{Definition of Mutual Information}
For the loss function, we choose InfoNCE\cite{tian2020contrastive} as the estimator. Since there are two view generators, for an input batch of $N$ graphs, the view generator will generate $2N$ augmented views. We regard the two augmented view representations $z_{i}$ and $z_{j}$ from the same input graph as positive view pairs, and $\operatorname{sim}(\cdot, \cdot)$ as cosine similarity function, $\tau$ is used as a temperature parameter constant. Then the mutual information $\operatorname{I}(i, j)$ of the positive view pair $(i, j)$ is expressed as
\begin{equation}\label{eq1}
\operatorname{I}(i, j)=\log \frac{\exp \left(\operatorname{sim}\left(z_{i}, z_{j}\right)/\tau\right)}{\sum_{k=1, k \neq i}^{2N} \exp \left(\operatorname{sim}\left(z_{i}, z_{k}\right)/\tau\right)}
\end{equation}

Due to the symmetric relationship between augmented views, the loss $\mathcal{L}_{\mathrm{max}}$ under this batch is expressed as:
\begin{equation}
\mathcal{L}_{\mathrm{max}}=-\frac{1}{2 N} \sum_{k=1}^N[\operatorname{I}(2 k-1,2 k)+\operatorname{I}(2 k, 2 k-1)]
\end{equation}

When minimizing the mutual information between the original graph and the views, we take the original graph as the anchor, and the loss can be further expressed as:

\begin{equation}\label{}
\mathcal{L}_{\mathrm{min}}=\frac{1}{2 N} \sum_{i=1}^{2 N}\log \frac{\exp \left(\operatorname{sim}\left(z_{i,1}, z_{i,2}\right)/\tau\right)}{\sum_{k=1, k \neq i}^{2N} \exp \left(\operatorname{sim}\left(z_{i,1}, z_{k,2}\right)/\tau\right)}
\end{equation}

\subsubsection{Joint Training Strategy}
For unsupervised joint training, at the beginning, we  pass a batch of data through the view generators with initialization parameters, and obtain the feature-evel and edge-level augmented view embeddings through the GNN encoder. At the same time, we pass this batch of data directly through the GNN to obtain the original data embedding.
By minimizing mutual information between augmented view embeddings and original data embedding, the view generators are updated adaptively. Note that the augmented view passed the GNN encoder during mutual information minimization has not been trained, after the preliminary update of the view generators, we use the same batch of data to further train the view generators. The data is passed through the view generators and GNN encoder to obtain view embeddings again.  We contrast feature-level and edge-level view embeddings by InfoMax to update the view generators and GNN encoder.

It is observed that only using augmented view representations for infomax may still not guarantee semantics and thus affect the performance of GNN. Therefore, we utilize random sampling of augmented view representations and original graph representation for Infomax, and guide the update of GNN and view generators.
The detail of training strategy is described in algorithm \ref{code:code1}.

\subsubsection{Analysis of Joint Training Strategy}
We further illustrate how our training strategy differs from other related work, we will compare the our joint training strategy wit AutoGCL\cite{yin2022autogcl} and AD-GCL\cite{suresh2021adversarial} . AutoGCL restricts the similarity of views to obtain sufficient contrastive variance, which cannot be achieved under unsupervised tasks, because it needs to be further trained in supervised way to guarantee the semantics of the raw data contained in the views. Therefore, the joint training strategy of  AutoGCL is based on semi-supervised learning rather than unsupervised learning. And our other constraint, that is, the minimization of mutual information, can fully adapt to the requirements of unsupervised tasks.
This produces sufficient augmentation variance because of the different augmentation angles chosen for each. It means that the GNN encoder needs to find out the semantic consistency of features and topology from the contrastive process of the two views.

For the training strategy of AD-GCL, in order to get a sufficiently lower bound of large semantic, AD-GCL requires a regularization term to guarantee the minimum semantic guarantee during mutual information minimization. 
Since regularization parameters need to be extensively evaluated, its optimization will still utilize weak label information in some downstream tasks.We do not introduce a regularization term to guarantee the semantic lower bound of the augmented view, instead, regularization can be completely replaced by our joint training strategy to guarantee the semantic lower bound, thus it is not required to exploit weak information in downstream tasks.

Specifically, when the view generators are jointly  trained, the mutual information between the original input data view and the two augmented views is minimized, so that the view generators can make augmentation as aggressive as possible, which is similar to the mutual information minimization process of AD-GCL. However,at the same time, the view generator is constrained by Infomax. This has the effect of constraining the minimization of mutual information and preserving the semantic lower bound. 

Although at the same time and space, InfoMax and mutual information minimization are not synchronously guiding the view generators, but from a logical point of view, InfoMax can still be considered as the limit of mutual information minimization, Because it forces the view generators to update in the opposite direction of mutual information minimization. 
The reason why it is not placed in the same space-time joint training, but in the way of joint alternate training, on the one hand, is that the contrastive samples of the two are different, on the other hand, the GNN encoder is only guided by InfoMax to capture consistency between view representations, so when minimizing mutual information, the GNN encoder has actually been further optimized by InfoMax, which is also similar to the adversarial strategy in AD-GCL.
\floatname{algorithm}{Algorithm}
\renewcommand{\thealgorithm}{3.2}
    \begin{algorithm} [H]
        \caption{Joint Training View Generators} 
        \label{code:code1}
        \begin{algorithmic}[1] 
            \State Initialize the parameters of the view generators $G_1$ and $G_2$ and the GNN encoder $f$;
            \For{training epochs} 
                \For{minibatch data $x$}     
                    \State get representations $z_0=f(x)$, $z_1=f(G_1(x))$, $z_2=f(G_2(x))$
                    \State $\mathcal{L}=\mathcal{L}_{\mathrm{min}}(z_0,z_1)$
                    \State update the parameters of $G_1$ to minimize $\mathcal{L}$
                    \State $\mathcal{L}=\mathcal{L}_{\mathrm{min}}(z_0,z_2)$
                    \State update the parameters of $G_2$ to minimize $\mathcal{L}$
                    \State get representations $z_1=f(G_1(x))$, $z_2=f(G_2(x))$
                    \State sample two representations $r_1$, $r_2$ from $\{z_0,z_1,z_2\}$
                    \State $\mathcal{L}=\mathcal{L}_{\mathrm{max}}(r_1,r_2)$
                    \State Update the parameters of $f$, one or two from $\{G_1, G_2\}$ to minimize $\mathcal{L}$
                \EndFor
            \EndFor
        \end{algorithmic}
    \end{algorithm}

\section{Experiments}
To evaluate the effectiveness and generality of our proposed model, we conducted extensive experiments on 
TU Dataset \cite{morris2020tudataset} to solve graph-level classification tasks. The TU dateset contains two categories of data, one is biochemical molecules and the other is the social networks data. Biosocial molecules contain additional node features. 
The statistics of the datasets are summarized in Table\ref{tab1}.
\begin{table}[H]
	\caption{ Summary of  TU Benchmark Dataset used for unsupervised  learning experiments.}
	\vspace{-3mm}
	\begin{center}
            \scalebox{0.65}{
		\begin{tabular}{c c c c c c}
			\toprule[2pt]
                {Datasets}&{Graphs} &  {Avg Nodes} & {Avg Edges} &  {Classes} &  {num features}
                \\
			\midrule[1pt]
                \multicolumn{6}{c}{Biochemical Molecules}\\
                \cmidrule(){1-6}
			{NCI1} & {4,110} &  {29.87} & {32.30} & {2}&  {37}
			\\
			{PROTEINS}   &  {1,113} & {39.06}&  {72.82}&  {2}&  {4}
			\\
			{DD}& {1,178}& {284.32}& {715.66}& {2} &  {89}
			\\
                 \cmidrule(){1-6}
                \multicolumn{6}{c}{Social Networks}\\
                \cmidrule(){1-6}
			{COLLAB}& {5,000} & {74.5}& {2457.78}& {3}&  {None}
			\\
			{REDDIT-BINARY}& {2,000} & {429.6}& {497.75}& {2}	& {None}	
			\\
                {REDDIT-MULTI-5K}&{4,999} & {508.8}&{594.87}& {5}& {None}
                \\
                {IMDB-BINARY}& {1,000} & {19.8}& {96.53}& {2}&  {None}
               \\
                {IMDB-MULTI}& {1,500} & {13.0}& {65.94}& {3}&{None}
			\vspace{1mm}
               \\
			\bottomrule[2pt]
		\end{tabular}
            }
	\end{center}
	\label{tab1}
\end{table}

\subsection{Experimental Settings}
For the unsupervised graph classification task, we use unlabeled data to train the model, then fix the model parameters and further use the labeled data to train the downstream classifier. Some of the work employs nonlinear classifiers, which are more powerful than linear classifiers in terms of their capabilities.In order to be fair, the downstream classifiers we use are unified as linear classifiers with the same settings: LibLinear\cite {2008LIBLINEAR} solver. We hope that the performance gain of any model should be attributed to the quality of the representations generated by the model.

For TU Dataset\cite{morris2020tudataset}, the batchsize is 32 and the 10-Fold evaluation is followed, the Accuracy is used as the standard protocol. All the experiments were conducted five times with different random seeds, the mean and standard deviation of the corresponding test indicators of each data set were reported.
For the encoder, all baselines and HAGCL use the same settings of 5-layer GIN\cite{xu2018powerful}, encoder fixed and not tuned. The GIN has an embedding dimension of 32, a pooling layer, no skip connection.
The optimization of HAGCL is based on Adam and the learning rate of encoder and view generator is set to the same 0.001.

\subsection{Performance Comparison}
We compare the performance of our proposed model with graph contrastive learning methods that are proposed in recent years, including InfoGraph\cite{2019InfoGraph}, GraphCL\cite{you2020graph}, AD-GCL\cite{suresh2021adversarial} and AutoGCL\cite{yin2022autogcl}.
Among them, AD-GCL-FIX uses fixed regularization hyperparameter to limit edge drop ratio when minimizing mutual information. AD-GCL uses the weak labels in the downstream tasks for extensive evaluation, and selects the optimal value from the optional regularization hyperparametric values.
Table \ref{tab2} shows the performance of different methods on all datasets.\begin{table*}[htbp]
	\caption{ Comparison with the existing methods for unsupervised learning.The \pmb{bold} numbers denote the best performance and the \textcolor{blue}{blue} numbers represent the second best performance.}
	\vspace{-3mm}
	\begin{center}
        \resizebox{\textwidth}{!}{
		\begin{tabular}{c c c c |c c c c c}
			\toprule[2pt]
			Dataset &  NCI1 &  PROTEINS & DD &  COLLAB &  RDT-B &  RDT-M5K &  IMDB-B&  IMDB-M
			\\
			\midrule[1pt]
			InfoGraph & 68.13±0.59 &   72.57±0.65 &  75.23±0.39& 70.35±0.64& 78.79±2.14& 51.11±0.55& 71.11±0.88& 48.66±0.67
			\vspace{1mm}
			\\
			GraphCL   &  68.54±0.55& 72.86 ± 1.01&  74.70±0.70&  71.26±0.55&   82.63±0.99&  53.05±0.40&  70.80±0.77& 48.49±0.63
			\vspace{1mm}
			\\
			AD-GCL-FIX& 69.67±0.51 & 73.59±0.65&  74.49±0.52&  \textcolor{blue}{73.32±0.61}& \textcolor{blue}{85.52±0.79}&  53.00±0.82& 71.57±1.01& 49.04±0.53
			\vspace{1mm}
			\\
			{AD-GCL}& {69.67±0.51}& \textcolor{blue}{73.81±0.46}& {75.10±0.39}& \textcolor{blue}{73.32±0.61}& \textcolor{blue}{85.52±0.79}& \textcolor{blue}{54.93±0.43}& \pmb{72.33±0.56}& \textcolor{blue}{49.89±0.66}
			\\
			{AutoGCL}& \pmb{74.61±0.86} & {73.32±0.47}&  \textcolor{blue}{76.29±0.35}& {69.74±0.63}&
            {84.21±0.62}&
            \pmb{56.01±0.41}&
            {71.14±0.45}&
            {48.44±0.81}
			\vspace{1mm}
			\\
			{HAGCL}& \textcolor{blue}{70.71±0.72} & \pmb{74.63±0.62}& \pmb{78.18±0.57}&  \pmb{73.64±0.46}& \pmb{86.80±0.72}& {54.62±0.61}& \textcolor{blue}{71.63±0.63}& \pmb{50.52±0.60}
			\vspace{1mm}
			\\
			\bottomrule[2pt]
		\end{tabular}
        }
	\end{center}
	\label{tab2}
\end{table*}
For Biochemical Moleculars, we uniformly use the default configuration of TD Datasets to provide support for all models. HAGCL achieved the best performance on the PROTEINS and DD datasets, and achieved a significant improvement on the DD dataset compared to previous work, and achieved sub-optimal performance on NCI1. We believe this is the result of the joint action of the view generator and joint training strategy. 

In order to better analyze the performance of each component and the joint training strategy, we further conducted extensive experiments on Social Networks.

\subsection{The Effectiveness of Joint Training Strategy.}
Since there are no node features in social networks, we set all node feature values to 1 and apply them uniformly to all works. Some baselines use more advanced preprocessing for node features, which can significantly improve accuracy. In order to make a fair comparison, we also unified them. At this time, the feature-level view generator can no longer play functions, so we separate this component and only use the edge-level view generator combined with the joint training strategy for training. 

When we compare our model with AD-GCL, both HAGCL and AD-GCL adopt edge drop, while AD-GCL adopts an adversarial strategy, and the view generator is only optimized when minimizing mutual information, and a regularization term is required to limit the edge drop ratio to ensure the semantic lower bound. HAGCL does not need to rely on labels and adopts a joint training mode. We expect to surpass AD-GCL-FIX with fixed regularization parameters in social network data and achieve results similar to AD-GCL.

Table \ref{tab2} demonstrates that HAGCL outperformes other methods on COLLAB, REDDIT-BINARY, REDDIT-MULTI-5K datasets, and achieved the second best result on IMDB-BINARY, surpassing AD-GCL-FIX. Furthermore, HAGCL outperformed AD-GCL on most of the datasets. The reason behind this success is that the optimal regularization hyperparameters of some data in AD-GCL are not within the range of optional values, and only the relatively best results can still be selected. In order to confirm the optimal regularization hyperparameters for specific data, more extensive work may be needed. Nevertheless, HAGCL can achieve similar or even better results than AD-GCL without relying on any labels or extra work, which proves the effectiveness of the joint training strategy.
\begin{figure}
    \centering
    \includegraphics[width=0.47\textwidth]{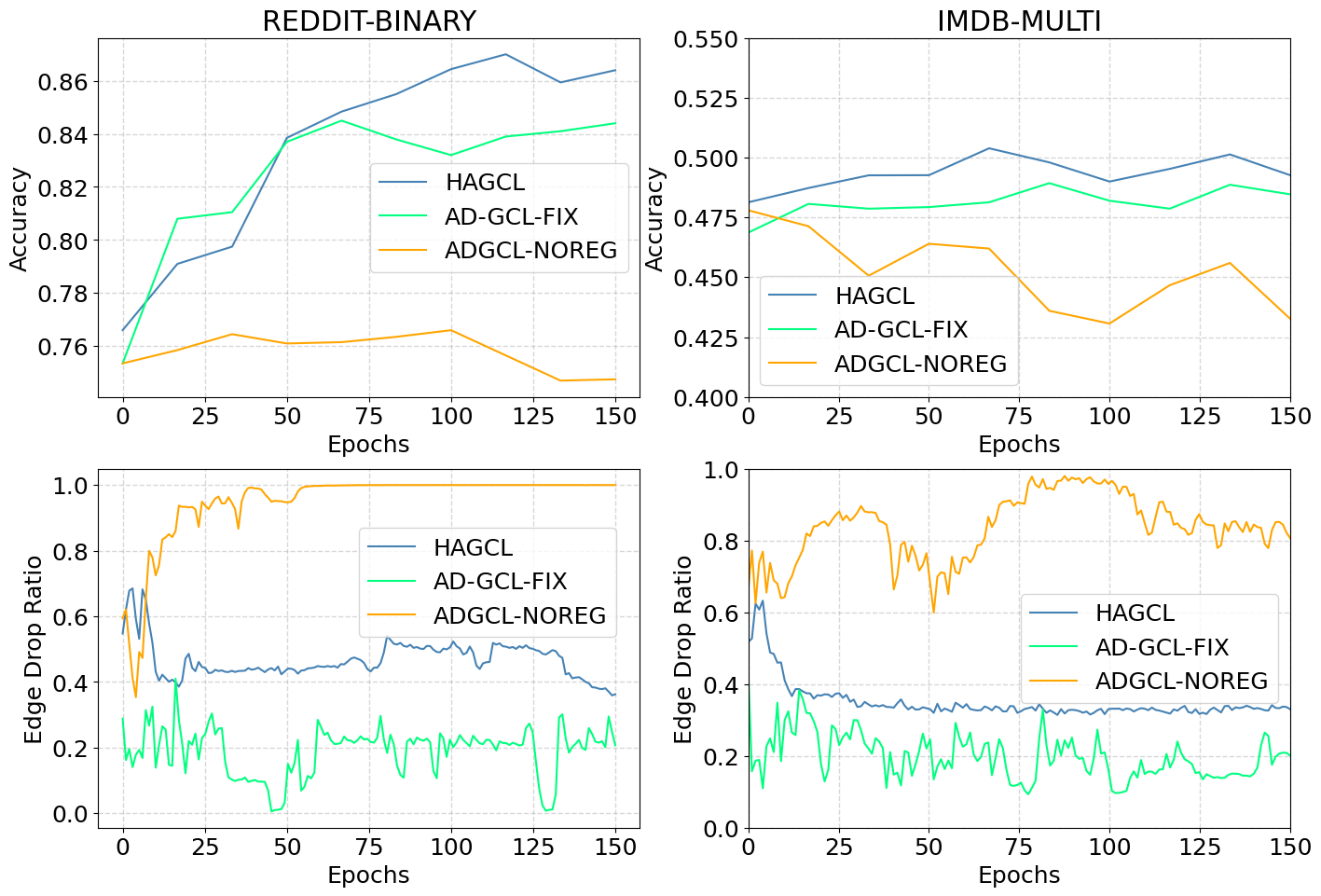}
    \caption{The performance changes of different methods on two datasets as the epoch increases
    }
    \label{fig3}
\end{figure}
We further analyze the edge drop ratio under joint training, and analyzed AD-GCL-FIX with fixed regularization terms and AD-GCL-NOREG without regularization terms. 

Figure \ref{fig3} visualizes the accuracy and edge drop ratio results of our model and other baseline methods in four datasets. We observe that the regularization parameter is crucial for AD-GCL. When the regularization term is removed, the edge drop ratio becomes too aggressive and even approaches to 1, resulting in a significant gap in accuracy compared to the other two methods. This is because the regularization term serves as a semantic lower bound. Moreover, compared to AD-GCL-FIX, HAGCL consistently achieves higher edge drop ratio and accuracy. This is due to the fact that the regularization term restricts the semantic lower bound, which is reflected in the fact that edge drop is always optimized within the constrained search space.

\subsection{The Effectiveness of View Generators}
\begin{figure}[!htp]
    \centering
    \includegraphics[width=0.47\textwidth]{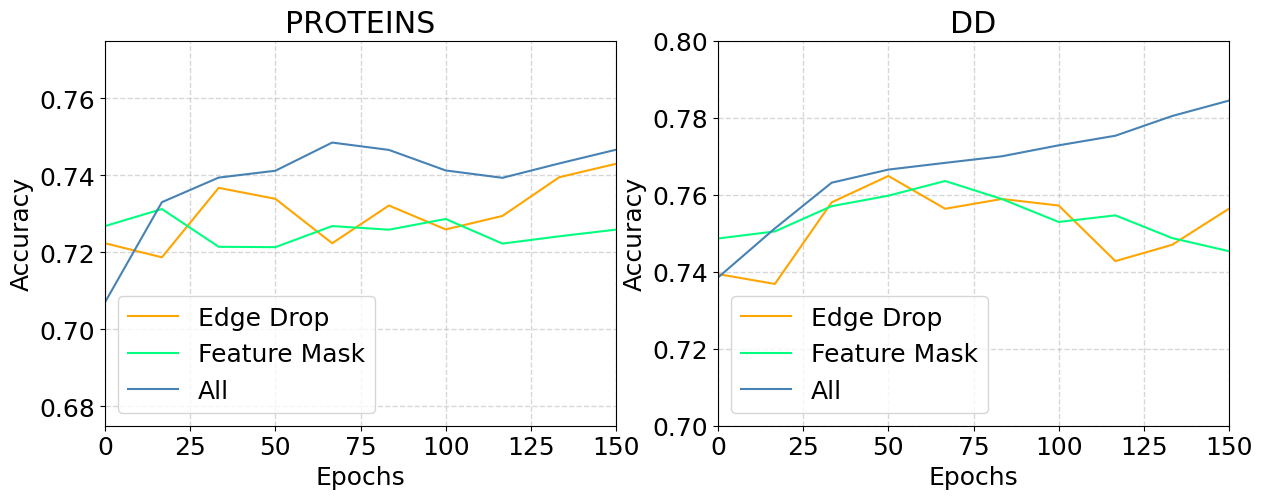}
    \caption{The ablation experiments for view generators
    }
    \label{fig4}
\end{figure}

We further analyze the effects of the two view generators in HAGCL, we conduct ablation experiments and anlyze experimental results on the Biochemical Molecules dataset. We design three variants of HAGCL, including:
\begin{itemize}
    \item  Edge Drop: It represents only using the edge-level view generator.
    \item Feature mask: It represents only using the feature-level view generator.
    \item All: It represents using two types view generators.
\end{itemize}
All the variants methods have the same settings as HAGCL except the differences mentioned above. Figure \ref{fig4} shows the accuracy results of the models under different epoches. It is obvious that The overall performance is always relatively better when both are involved. We believe that this is because the GNN encoder can learn semantic commonalities from different perspectives of the views.

Due to the involvement of a single view generator, when using Infomax, it operates only on the original graph and a single angle augmented graph, thus GNN can only learn graph semantics from a single angle. However, when they work together, Infomax samples on the original graph and augmented graphs from different perspectives, which not only enables GNN to learn semantic commonality under multiple scale augmentation, but also allows for more comprehensive learning of semantics from each perspective.

\section{Conclusion}
In this paper, we propose an effective graph contrastive representation learning framework with with hybrid augmentation perspectives. For unsupervised tasks, we propose an improved joint training strategy that utilizes Infomax and mutual information minimization to jointly optimize data augmentation, which removes redundancy of augmented views with semantics preserving. In particular, our approach can further support the GNN encoder in learning semantic commonalities under different augmentation angles. It is worth mentioning that the proposed method eliminates the need for setting augmentation hyperparameters or regularization terms and does not rely on weak information from downstream tasks, achieving true automation of augmentation in unsupervised tasks. Extensive experiments demonstrate the superiority of our approach over state-the-art methods in unsupervised graph representation learning.

\bibliographystyle{kr}
\bibliography{article}

\end{document}